\PassOptionsToPackage{dvipsnames}{xcolor}
\documentclass[runningheads]{llncs}
\usepackage{amsmath} 
\usepackage{amsfonts} 
\usepackage[T1]{fontenc}
\usepackage{graphicx}
\usepackage{hyperref}               
\usepackage{color}

\usepackage{array} 
\usepackage{subcaption} 
\usepackage{booktabs}
\usepackage{colortbl}
\usepackage{multirow}

\usepackage{pifont} 
\usepackage{todonotes}
\usepackage[dvipsnames]{xcolor}

\newcolumntype{P}[1]{>{\centering\arraybackslash}p{#1}}
\newcolumntype{M}[1]{>{\centering\arraybackslash}m{#1}}

\newcommand{\OurMethod}{TEAR}

\definecolor{VisionBlue}{RGB}{218,232,252}
\definecolor{LanguageOrange}{RGB}{255,230,204}
\definecolor{ModelGreen}{RGB}{213,232,212}
\definecolor{FunctionPurple}{RGB}{225,213,231}
\definecolor{ParamRed}{RGB}{248,206,204}
\definecolor{ActionClass}{RGB}{250, 233, 226}
\definecolor{ActionDescriptor}{RGB}{250,226,231}
\definecolor{PredictionPurple}{RGB}{233,226,250}

\newcommand{\inlineColorbox}[2]{\begingroup\setlength{\fboxsep}{1pt}\colorbox{#1}{\hspace*{2pt}\vphantom{Ay}#2\hspace*{2pt}}\endgroup}

\newcommand{\rebuttal}[1]{\textcolor{black}{#1}} 

\begin{document}


\title{Text-Enhanced Zero-Shot Action Recognition: A training-free approach}
\titlerunning{Text-Enhanced Zero-Shot Action Recognition}
\authorrunning{Bosetti et al.}

\author{
Massimo Bosetti\inst{1}  \and
Shibingfeng Zhang\inst{3}  \and
Bendetta Liberatori\inst{1}  \and 
Giacomo Zara \inst{1} \and
Elisa Ricci\inst{1,2} \and
Paolo Rota\inst{1}}
%

%
\institute{University of Trento \\
\and
Fondazione Bruno Kessler (FBK) \\
\and
University of Bologna
}
\maketitle              

\begin{abstract}
Vision-language models (VLMs) have demonstrated remarkable performance across various visual tasks, leveraging joint learning of visual and textual representations. While these models excel in zero-shot image tasks, their application to zero-shot video action recognition (ZSVAR) remains challenging due to the dynamic and temporal nature of actions. Existing methods for ZS-VAR typically require extensive training on specific datasets, which can be resource-intensive and may introduce domain biases. In this work, we propose \textbf{T}ext-\textbf{E}nhanced \textbf{A}ction \textbf{R}ecognition (\textbf{\OurMethod}), a simple approach to ZS-VAR that is training-free and does not require the availability of training data or extensive computational resources. Drawing inspiration from recent findings in vision and language literature, we utilize action descriptors for decomposition and contextual information to enhance zero-shot action recognition. 
Through experiments on UCF101, HMDB51, and Kinetics-600 datasets, we showcase the effectiveness and applicability of our proposed approach in addressing the challenges of ZS-VAR.
\footnote{The code will be released later at \url{https://github.com/MaXDL4Phys/tear}}

\keywords{Action Recognition  \and Zero-shot Transfer \and Vision and Language}

\end{abstract}

\section{Introduction}
Multimodal vision-language models (VLMs)~\cite{radford2021learning,jia2021scaling} have demonstrated outstanding performance across diverse visual tasks.  These models undergo pre-training on large-scale datasets, aiming to jointly learn representations for images and text.
Benefiting from textual representations, VLMs have exhibited impressive zero-shot capabilities, \textit{i.e.}, ability to generalize to a novel set of unseen classes on a handful of tasks, such as image classification~\cite{zhang2021tip}, object detection~\cite{gu2022openvocabulary,Hanoona2022Bridging} and segmentation~\cite{rao2022denseclip}. However, despite the zero-shot transfer results achieved on image tasks, these models still struggle when applied zero-shot to videos without proper fine-tuning~\cite{ju2022prompting,nag2022zero,yan2023unloc}. 
Understanding actions in video streams is inherently more challenging than recognizing static elements in images. For instance, while identifying an object in an image may be straightforward, grasping intricate actions, such as dancing, involves understanding dynamic movements and temporal context, adding complexity to the task.
This characteristic makes video action recognition, which finds real-world applications in various fields~\cite{deng2023large} like autonomous driving, sports analysis, and entertainment, typically more challenging than the image counterpart.

Recognizing actions in videos through zero-shot video action recognition (ZS-VAR) using VLMs can be challenging due to the associated temporal dynamics and complexities. Additional training is often required to capture these factors. Recent ZS-VAR methods have shown satisfactory results but require extensive training on appropriate datasets to achieve such performance.~\cite{lin2023match,wang2021actionclip,ni2022expanding,ju2022prompting}.
While effective, these approaches have several drawbacks.
Primarily, the training process can be time-consuming and resource-intensive. 
Additionally, fine-tuning task-specific datasets may introduce biases into the system, limiting its generalizability across different datasets~\cite{liberatori2024testtime}.
Furthermore, introducing new parameters can increase the computational cost of model deployment and inference, adding to the complexity of these approaches in real-world scenarios.

These motivations prompt us to explore an alternative approach to ZS-VAR that is training-free and does not require the availability of training data or extensive computational resources. 
One recently highlighted problem of VLMs is that they may not encode sufficient knowledge of verbs, which are crucial for understanding and recognizing actions in videos~\cite{momeni2023verbs,park2022exposing,wang2024paxion}. Additionally, research has shown that incorporating contextual information in textual prompts can enhance the performance of VLMs in various downstream tasks~\cite{menon2023visual,an2023more}.
Drawing inspiration from these recent findings, we aim to leverage 
the decomposition of actions and the introduction of contextual information to improve zero-shot action recognition without further training.

We propose \OurMethod, which stands for Text-Enhanced Zero-Shot Action Recognition, as a training-free approach for ZS-VAR. We leverage a VLM pre-trained solely on image data, abstaining from fine-tuning it on video data. 
Our approach unfolds in two primary steps: first, the generation of action descriptors employing a large language model (LLM); second, zero-shot prediction facilitated by the generated textual descriptors.
We evaluate the proposed approach on three standard benchmarks, \textit{i.e.}, UCF101 \cite{Soomro2012-ah}, HMDB51~\cite{Kuehne2011-uk}, and Kinetics-600~\cite{Carreira2018-ly}.

Our contributions can be summarized as follows:
\begin{enumerate}
 \item We propose \OurMethod, Text-Enhanced Action Recognition, the first method addressing zero-shot video action recognition in a training-free manner. Our approach does not rely on the availability of training data or require significant computational resources. This contribution makes ZS-VAR more accessible and practical for real-world applications.
\item By decomposing action labels into sequential observable steps and providing visually related descriptions, our approach enables better understanding and recognition of actions in videos. We demonstrate how leveraging decomposition and description benefits the zero-shot action recognition task.
\item We empirically show the capabilities of the proposed method on three datasets, \textit{i.e.}, UCF101~\cite{Soomro2012-ah}, HMDB51~\cite{Kuehne2011-uk}, and Kinetics-600~\cite{Carreira2018-ly}, achieving results that are competitive with training-based approaches. 

\end{enumerate}

\section{Related Work}

\subsubsection{Vision-language models.}\label{sss:vlmodel}
Vision-language models (VLMs), such as CLIP~\cite{radford2021learning}, have been developed to learn joint visual-text embedding spaces through pre-training on large-scale datasets of web-crawled image-text pairs.
They have showcased outstanding performance across various downstream tasks, particularly in the image domain, with notable zero-shot capabilities~\cite{jia2021scaling}. These models have been recently extended to the video domain, where tasks are typically more challenging due to the additional temporal dimension.
Recent works achieve this by incorporating additional learnable components for spatiotemporal
modeling, including self-attention layers, textual or vision prompts, or dedicated visual decoders, demonstrating improvements in video-related tasks~\cite{wang2021actionclip,ju2022prompting}.
However, their adaptation to zero-shot settings still necessitates further development, and the results currently lag significantly behind those achieved in tasks related to image processing.
Moreover, by introducing new parameters, these methods necessitate additional training and the availability of large-scale training data. This dependence makes the adaptation resource-intensive and can further introduce domain bias, limiting zero-shot transfer on unseen classes.

\subsubsection{Zero-shot action recognition.}\label{sss:zsar}
Zero-shot action recognition consists of identifying actions in videos from a closed set of action classes not encountered during the model's training phase. 
Early work~\cite{liu2011recognizing,zellerschoi2017zero} proposed to represent actions by sets of manually defined high-level semantic concepts, \textit{i.e.}, attributes, and show that this can be used to recognize action categories that have never been seen before. This advancement represented a step toward more explicit, semantics-driven solutions, as opposed to the modeling of input sequences in latent spaces~\cite{Doshi_2023_CVPR,transport,10115032,10093084}.
Another line of work~\cite{brattoli2020rethinking,mandal2019out,qin2017zero,shao2020temporal} uses word embedding of action names as semantic representation. Our work differs from these due to the idea of the language modality alone being the key for generalizing to new tasks and categories in a specifically video-oriented fashion.
 
\subsubsection{Vision-language for action recognition}
Previous works have explored the potential of leveraging the newly advanced VLMs, such as CLIP~\cite{radford2021learning}, to enhance recognition capabilities with textually conveyed semantics~\cite{menon2023visual,pratt2023does,roth2023waffling}.
These works, however, address the more generic task of image-based recognition without employing text-oriented solutions tailored for videos. On the other hand, the video field has been investigated in many subsequent works based on video captioning~\cite{tellme} and improved textual descriptors~\cite{wang2021actionclip,ni2022expanding,zhu2023orthogonal,yang2024epk,ranasinghe2024language}.
\rebuttal{Our work is more closely aligned with methods using LLMs~\cite{lin2023match,huang2024froster,ranasinghe2024language}.
MAXI~\cite{lin2023match} adapts a VLM for zero-shot action recognition using only unlabelled videos, composing a text bag for each unlabelled video using a captioning model and an LLM.  
FROSTER~\cite{huang2024froster} tackles open-vocabulary action recognition and uses an LLM as a form of text augmentation at training time to mitigate the distribution shift between CLIP's pre-training captions and template-embedded action names. 
}

However, we move a step further towards specifically video-oriented solutions by employing text-based augmentations to the label space that explicitly leverage the temporal and sequential nature of video data, such as the decomposition of action into sequential sub-actions.

\section{Method}\label{sec:method}
\begin{figure}
     \centering
     \includegraphics[width=1\linewidth]{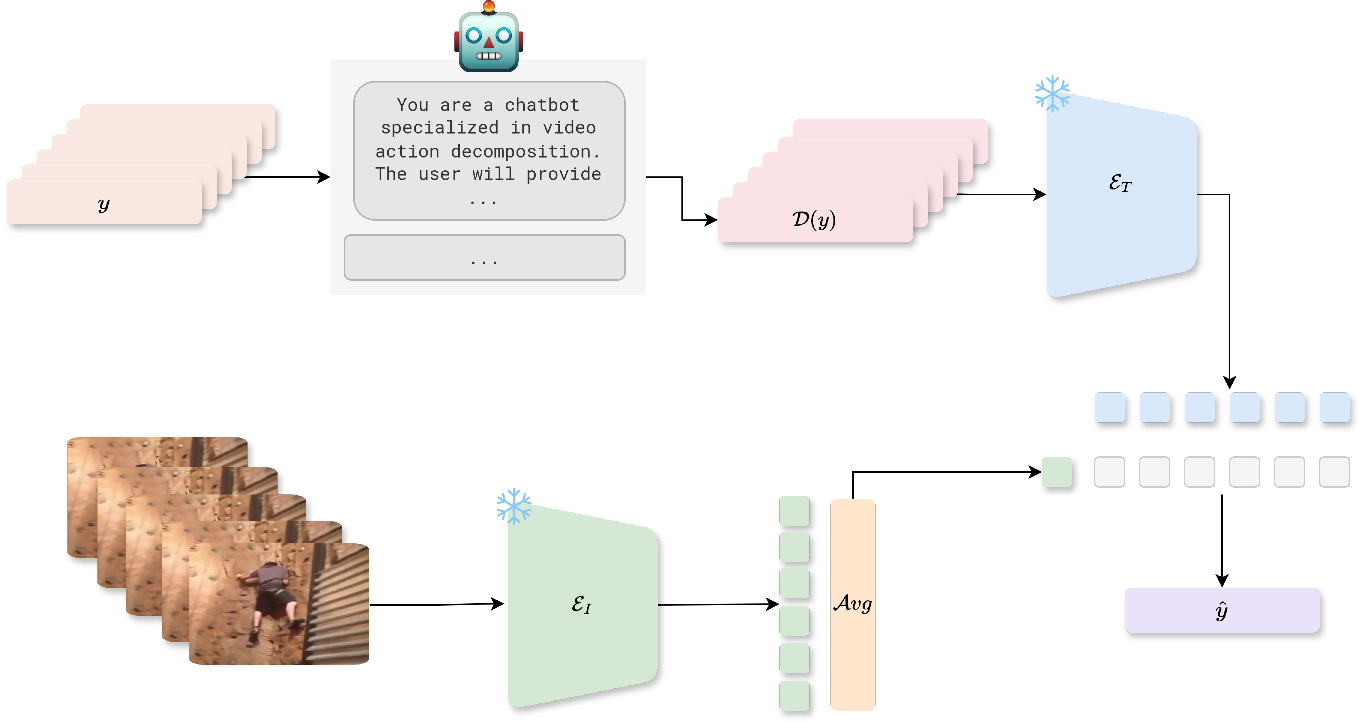}
     \caption{\textbf{Overview of the proposed method.} \OurMethod~addresses the task of zero-shot action recognition. First, for every \inlineColorbox{ActionClass}{action class label} $y$, we generate a set of \inlineColorbox{ActionDescriptor}{action textual descriptors} $\mathcal{D}(y)$ by querying an LLM. Then we compute the \inlineColorbox{VisionBlue}{textual} and \inlineColorbox{ModelGreen}{visual} embeddings, keeping both the image and text encoders frozen (\raisebox{-1.3mm}{\includegraphics[height=4.3mm]{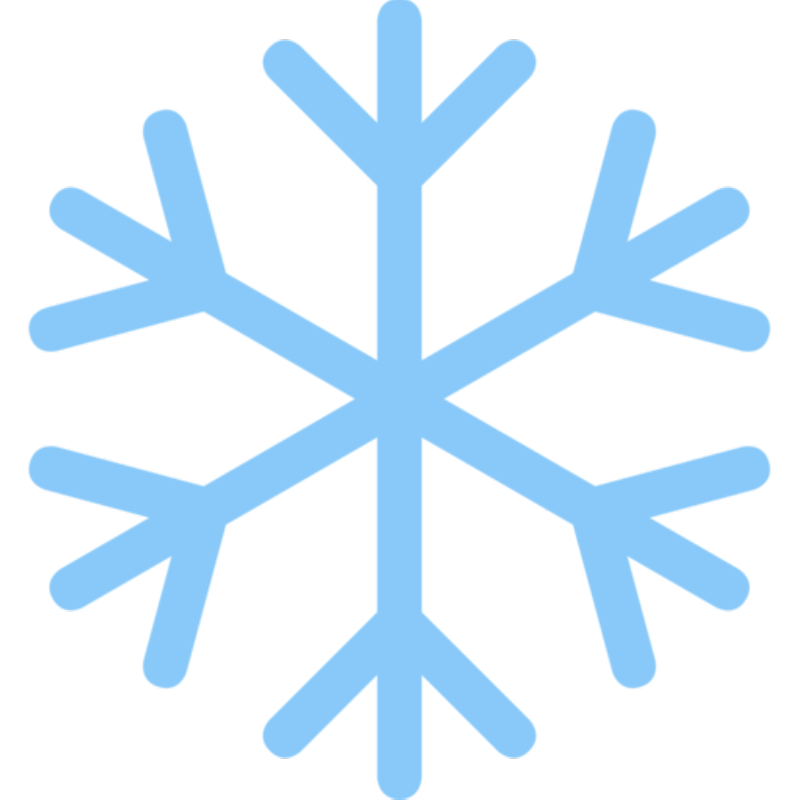}}). Lastly, the \inlineColorbox{PredictionPurple}{final prediction} is obtained by computing the similarity between the textual embeddings and the averaged visual embeddings. }
     \label{fig:tear_method}
 \end{figure}

Most prior works in zero-shot video action recognition have focused on adapting image-based VLMs through additional training, necessitating video data availability. 
In this work, we propose to leverage a language-driven manipulation of action labels and demonstrate that it enables effective action recognition without the need for further training, thereby achieving zero-shot performance. Our proposed method \OurMethod~directly addresses zero-shot action recognition at inference time in a remarkably simple but efficient way, as illustrated in Fig.~\ref{fig:tear_method}. 

Formally, given a video $\mathcal{V}$ and a pre-defined set of action classes $\mathcal{C}$, our goal is to classify the action present in the video. We achieve this with an image-based VLM model and an LLM, without necessitating tailored fine-tuning on video data. \OurMethod~employs a pre-trained CLIP~\cite{radford2021learning} as the VLM and GPT-3.5~\cite{brown2020language} for the LLM. 
The method consists of two main steps: firstly, generating action textual descriptors using a large language model (LLM), and secondly, facilitating zero-shot prediction through these descriptors. We provide detailed explanations of these steps in the following.

\subsection{Action descriptors generation}\label{subsec:descriptors}
It has been shown that VLMs often struggle with verbs due to their strong object and noun bias~\cite{momeni2023verbs}. Our key insight is that an action is more than just the verb; the surrounding context and objects can describe it, the different steps needed to perform it and additional visual cues. 
For each category $y\in \mathcal{C}$ we construct a set of textual descriptors $\mathcal{D}(y)$, considering the following descriptors:

\begin{itemize}
    \item{\bf Class}: the action label $y$ in the original format. 
    \item{\bf Decomposition}: a list of sub-actions. 
    Specifically, we break down the action into three consecutive stages, capturing it across different temporal phases.
    \item{\bf Description}: an elaborate semantic description of the action.
    \item{\bf Context}: a textual descriptor encompassing two distinct types of information pertinent to the action. One is the overall context, highlighting visual features likely to be observed in a video portraying the action. The other is a list of objects likely to participate in the action. 
    \item {\bf Combinations}: a combination of all the previously listed ones.  
\end{itemize}

Crafting textual descriptors for classes manually becomes increasingly impractical as the number of datasets and classes grows, rendering it infeasible. For this reason, we propose to automatically construct this set by prompting a large language model, such as GPT-3.5~\cite{brown2020language}, with multiple queries. We design a query for each one of the textual descriptors, as reported in Tab.~\ref{tab:descriptors}. 
An example of the obtained descriptors for the action \textit{snowboarding} is reported in Tab.~\ref{tab:snowboarding_example}.

\begin{figure}[htbp]
    \centering
    \begin{subfigure}{\textwidth}
    \caption{\texttt{"making pizza"}}

                \includegraphics[width= \textwidth]{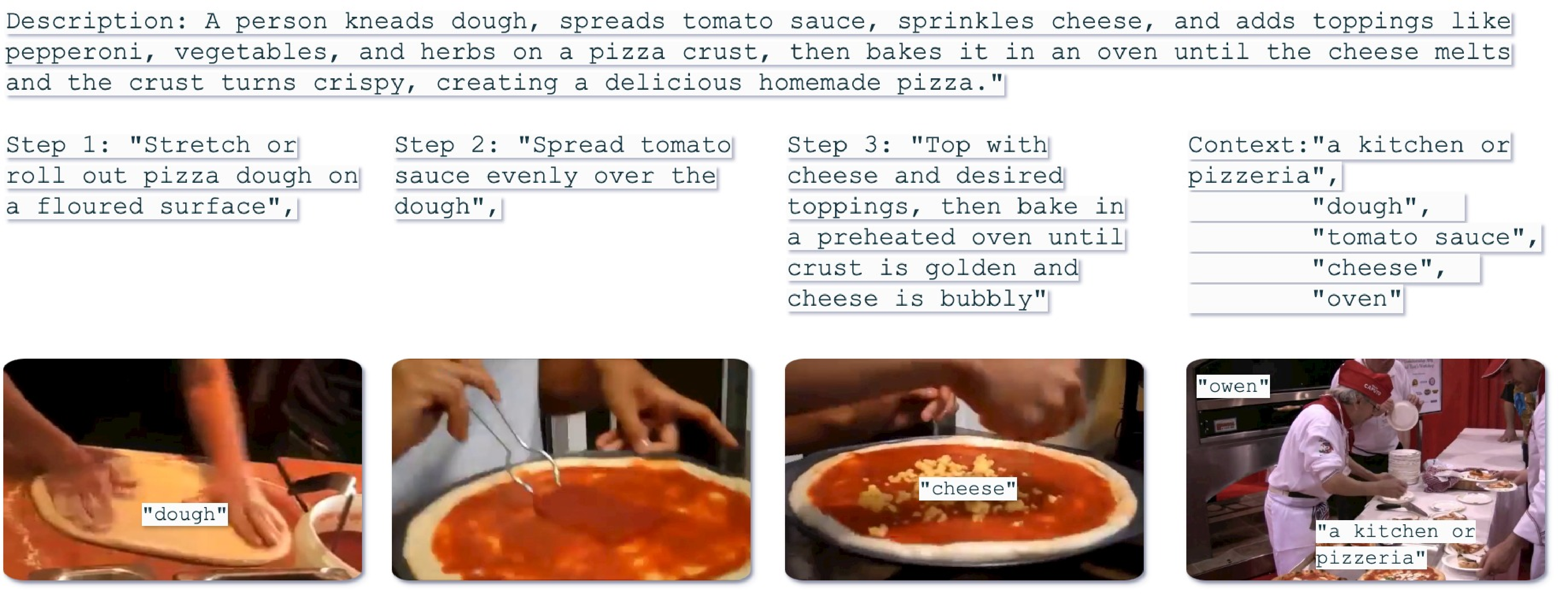}
        \label{fig:sub1}
    \end{subfigure}
    
    \begin{subfigure}{\textwidth}
        \caption{\texttt{"bowling"}}

         \includegraphics[width= \textwidth]{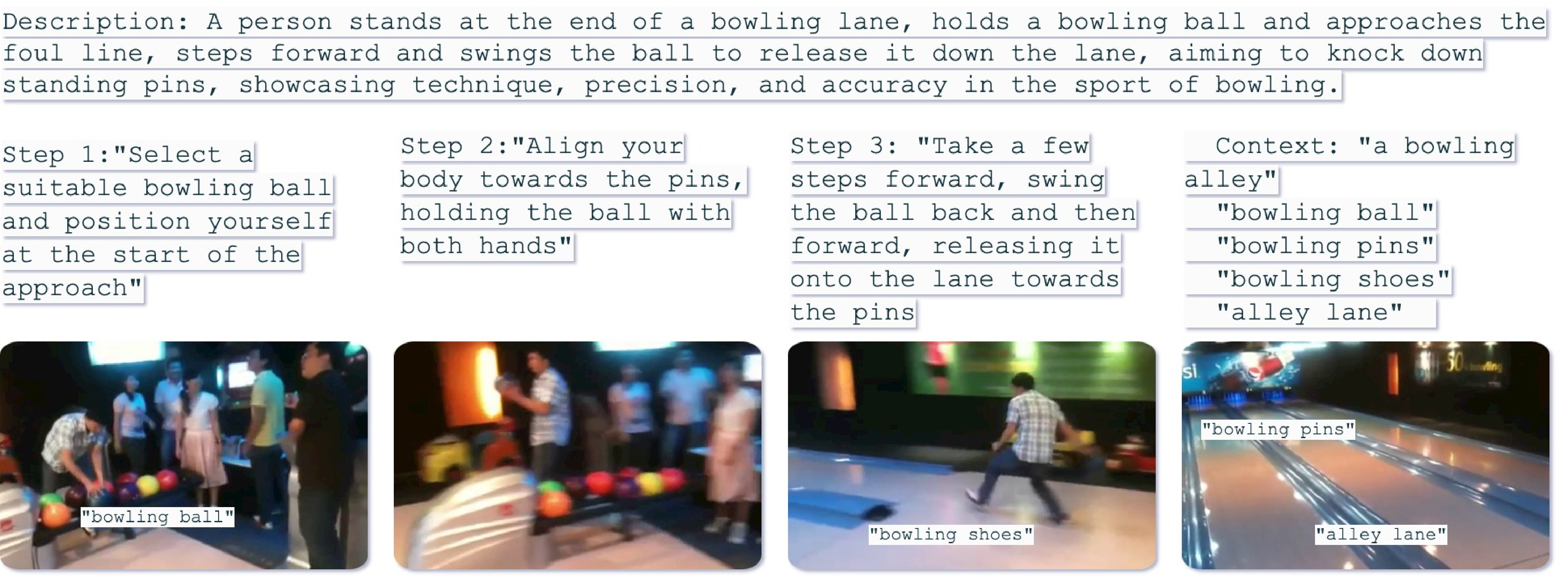}
        \label{fig:sub2}
    \end{subfigure}
    \begin{subfigure}{\textwidth}
        \caption{\texttt{"looking phone"}}

        \includegraphics[width= \textwidth]{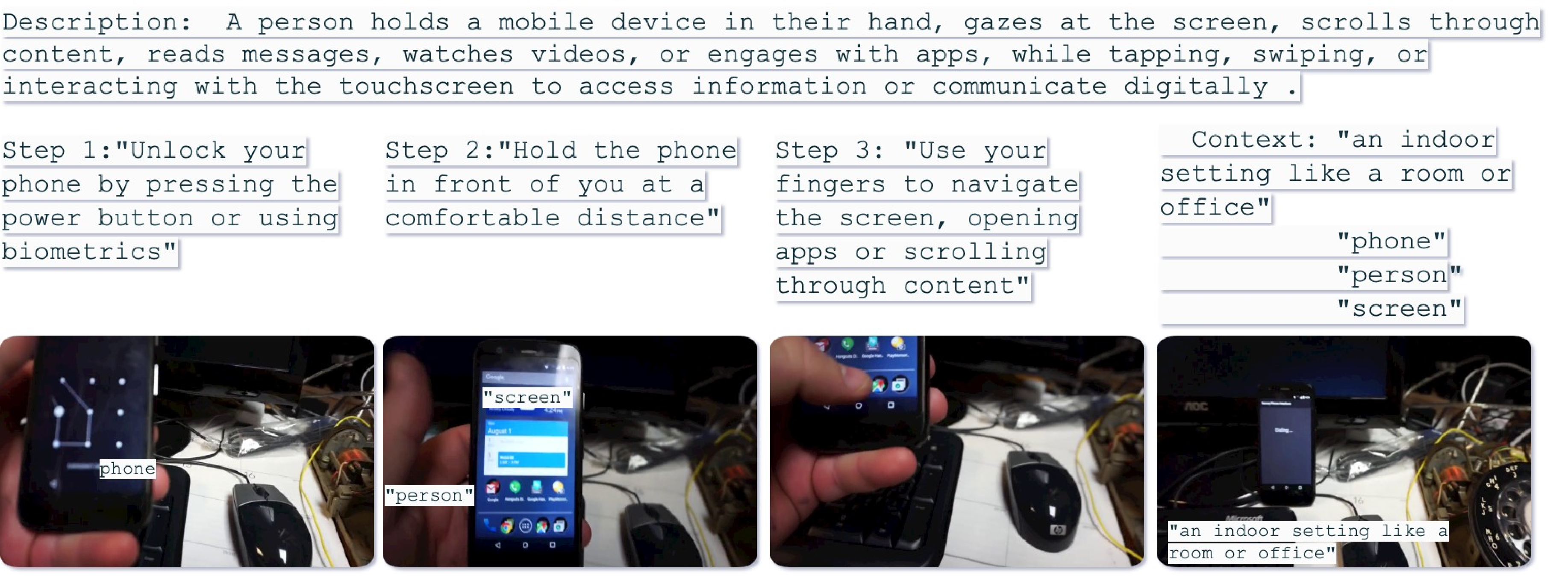}
        \label{fig:sub3}
    \end{subfigure}
        \caption{\textbf{Examples of descriptors matching visual cues in test videos.} We show descriptors generated for four videos of Kinetics-600. We show four frames for each video and highlight the matching with the decomposition, description, and context. For each video, the label above represents the ground truth label.}
    \label{fig:overall}
\end{figure}

\begin{figure}
    \begin{subfigure}{\textwidth}
        \caption*{\texttt{"kissing"}}
        \includegraphics[width=\textwidth]{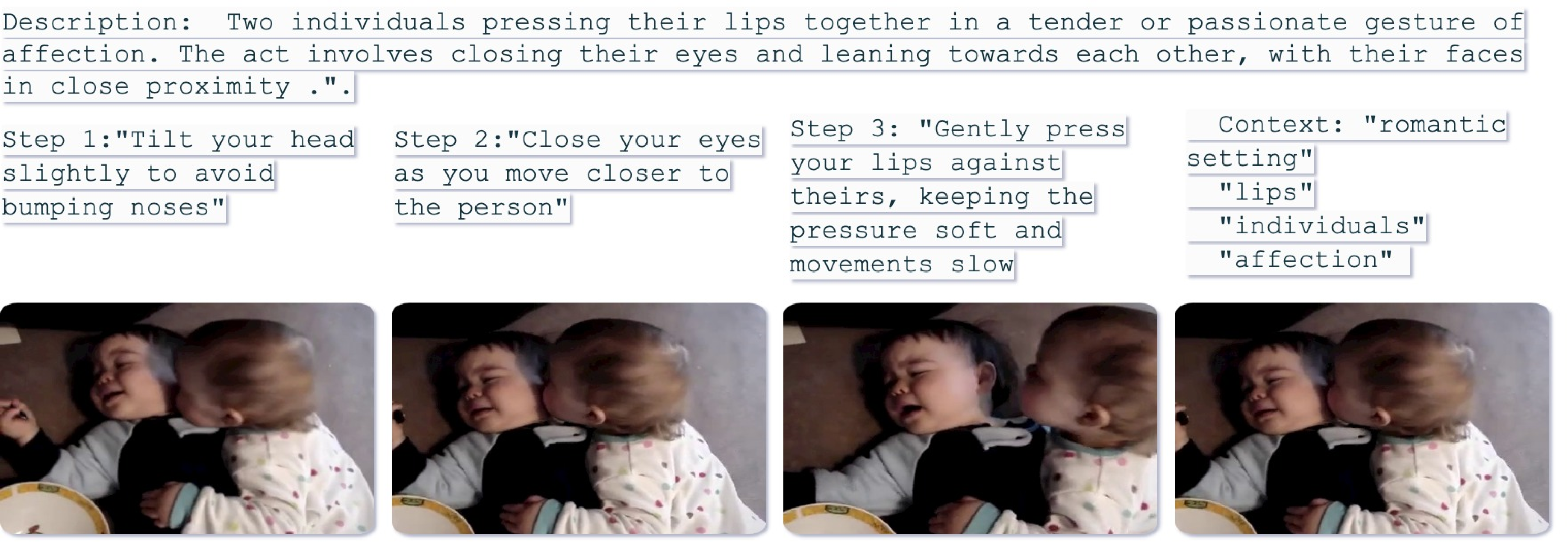}
        
        \label{fig:sub4}
        \end{subfigure}
        \caption{\textbf{Example of descriptors that do not match visual cues in test videos.} We show descriptors generated for one video of Kinetics-600 of the class \texttt{kissing}. We show four frames from the video and highlight the matching with the decomposition, description, and context. For this sample, the textual descriptors do not match the visual cues in the video. \rebuttal{Further qualitative analyses are available in the supplementary material.}
}
\label{fig:badcase}
\end{figure}

Visual inspection of the obtained $\mathcal{D}(y)$ against actual video content of the corresponding action class $y$ confirms the descriptors' relevance.
In Fig.~\ref{fig:overall} and~\ref{fig:badcase}, we illustrate a few examples of action labels, the generated descriptors, and four frames from a video of the same ground truth action.  In particular, Fig.~\ref{fig:sub1}, Fig.~\ref{fig:sub2} and  Fig.~\ref{fig:sub3} show that: i) the decomposition into steps corresponds to the sub-actions present in the video, describing the whole event in a set of more atomic actions, ii) the description is aligned with the general video content and, iii) the context and objects tags can be found in the video. This approach may result in failure cases when the textual descriptors do not accurately capture specific nuances. In Fig.~\ref{fig:badcase}, for example, the obtained descriptors depict a kissing action in a romantic setting, while a video labeled with the same action portrays a friendly interaction between babies.

Motivated by previous research demonstrating the efficacy of prompt templates~\cite{radford2021learning}, we incorporate a diverse set of templates, listed in supplementary material, into our approach. Specifically, we encapsulate all the obtained textual descriptors with the templates. 
Moreover, prepending the action class for each descriptor typically enhances performance. We attribute this to the fact that omitting the action class altogether and relying solely on the generated descriptors can result in a loss of information. This approach ensures that the generated descriptions maintain relevance and specificity.


\begin{table}[!ht]
\centering
\resizebox{1\textwidth}{!}{
\setlength{\tabcolsep}{2.5pt} 
\begin{tabular}{p{2.5cm}p{10cm}} 
\textbf{Descriptor} & \textbf{Query}  \\ 
\midrule
\cellcolor{FunctionPurple}
\textbf{Decomposition} & You are a chatbot specialised in video action decomposition. The user will provide you with an action and you will have to decompose it into three sequential observable steps. The steps must strictly be three. You must strictly provide each response as a python list, e.g., [’action1’, ’action2’, action3’]. Omit any kind of introduction, the response must only contain the three actions. Comply strictly to the template. Do not ask for any clarification, just give your best answer. It is for a school project, so it’s very important. It is also very important your response is in the form of a python list. \\ \noalign{\vskip1mm}
\midrule
\cellcolor{LanguageOrange}
\textbf{Description} & You are a chatbot specialised in video action description. The user will provide you with an action and you will have to describe the action by providing only visually related information. You must strictly provide each response as a Python string. The description should be succinct and general. Omit any kind of introduction. Comply strictly to the template. Do not ask for any clarification, just give your best answer. Following is an example. Action label: typing. Description: Typing normally involve a person and a device with keyboard. When typing, the individual positions their fingers over the keyboard. \\ \noalign{\vskip1mm}
\midrule
\cellcolor{VisionBlue}
\textbf{Context} & You are a chatbot specialised in video understanding. The user will provide you with the name of an action, and you will have to provide two specific pieces of information about that action. The first one is the context, which consists of any visually relevant feature that may be expected to appear in a video portraying that action. The second one consists of a lists of objects that may involved in the action. You must strictly provide each response as a python dictionary, e.g., ’context’: ’a person’, ’objects’: [’person’]. Omit any kind of introduction, the response must only contain the two pieces of information. Comply strictly to the template. Do not ask for any clarification, just give your best answer. It is for a school project, so it’s very important. It is also very important your response is in the form of a python dictionary. \\ \noalign{\vskip2mm} 
\end{tabular}
} %
\caption{\textbf{Queries used for action description generation.} We show the prompts used to query the LLM for each textual descriptor generation.}
\label{tab:descriptors}
\end{table}

\begin{table}[!htb]
\centering
\begingroup

\begin{tabular}{p{0.26\linewidth} p{0.64\linewidth}}
\multirow{-2}{*}{} \textbf{Descriptor} & \multirow{-2}{*}{} \textbf{Content}     \\

\toprule
        \textbf{Class}              &   \texttt{"snowboarding"}                                     \\
\midrule 
\cellcolor{FunctionPurple}
       \textbf{Decomposition}     &   \texttt{"Strap your feet securely onto the snowboard bindings",
                                    "Lean forward to initiate movement down the slope",
                                    "Use heel-to-toe shifts in weight to steer and balance as you descend"}                                          \\ 
\midrule 
\cellcolor{LanguageOrange}
        \textbf{Description}       &   \texttt{"A person sliding down a snow-covered slope on a single board attached to their feet, making turns and jumps while maintaining                               balance."}                                          \\

\midrule 
\cellcolor{VisionBlue}
        \textbf{Context}           &   \texttt{"snow-covered mountain slope or snow park", "snowboard", "snow boots", "helmet"}                                          \\ 
\midrule 
\cellcolor{ModelGreen}
\textbf{Combination}        &   \texttt{"snowboarding",
                                    "Strap your feet securely onto the snowboard bindings",
                                    "Lean forward to initiate movement down the slope",
                                    "Use heel-to-toe shifts in weight to steer and balance as you descend", "A person sliding down a snow-covered slope on a single board attached to their feet, making turns and jumps while maintaining balance.", "snow-covered mountain slope or snow park", "snowboard", "snow boots", "helmet"}                                         \\                    
\bottomrule
\noalign{\vskip2mm} 
\end{tabular}
\caption{\textbf{Example of generated action description}. We show an example for the specific action of \textit{snowboarding}.}
\label{tab:snowboarding_example}
\endgroup
\end{table}

\subsection{Zero-shot recognition with action descriptors}

\OurMethod~operates in the following straightforward manner to generate the final inference based on the previously discussed textual descriptors provided by the language model. The key components are a pre-trained VLM, consisting of an image encoder $\mathcal{E}_I$ and a text encoder $\mathcal{E}_T$. 
Given a test video $\mathcal{V}$, first, we sample $N$ frames uniformly along the whole duration of the video and represent it as a set of frames as $\mathcal{V}=\{x_i\}_{i=1}^N$. Then we compute a compact representation from $\mathcal{V}$ as the average of its $N$ frames’ latent representations, extracted with the vision encoder:

\begin{equation}
\bar{\mathcal{V}} = \frac{1}{N} \sum\limits_{i=1}^{N} \mathcal{E}_I\left(x_i\right)
\end{equation}

Then we compute a textual representation for each class $y_j\in\mathcal{C}$, by encoding the textual descriptors $\mathcal{D}=\{d_j\}_{j=1}^M$ with the text encoder and averaging them:

\begin{equation}
z_j = \frac{1}{M} \sum\limits_{i=1}^{M} \mathcal{E}_T\left(d_i\left(y_j\right)\right)
\end{equation}

\noindent where $M$ is not fixed and depends on the category $y_j$.

Lastly, our model selects the action with the highest cosine similarity to the compact video representation, allowing it to make the final predictions:

\begin{equation}\label{eq:prediction}
\hat{y} = \underset{j \in |\mathcal{C}|}{\mathrm{argmax}} \, \left( \frac{z_j \cdot \bar{\mathcal{V}}}{\|z_j\| \cdot \|\bar{\mathcal{V}}\|} \right)
\end{equation}

\section{Experiment Results}

\subsection{Datasets and Metrics}
We conduct experiments with three popular video action recognition datasets: UCF101~\cite{Soomro2012-ah}, HMDB51~\cite{Kuehne2011-uk}, and Kinetics-600 (K600)~\cite{Carreira2018-ly}. These datasets are frequently used to evaluate zero-shot action recognition. We report the standard evaluation metrics of Top1/Top5 accuracy. To ensure our experiments are comparable to previous studies, we adopt the same protocol of previous works~\cite{lin2023match,rasheed2022fine}. 

The HMDB51 dataset~\cite{Kuehne2011-uk} contains approximately 7,000 manually annotated videos of human motion sourced from various platforms, including films and YouTube. Each video is categorized under one of 51 action labels, with at least 101 videos per label. The average duration of each video is 3.2 seconds.

\noindent\textbf{UCF101.}
The UCF101 dataset~\cite{Soomro2012-ah} consists of 13,320 videos derived from various online platforms and categorized into 101 action classes. These classes encompass a wide range of human activities and are organized into five broad categories: Human Object Interaction, Body-Motion Only, Human-Human Interaction, Playing Musical Instruments, and Sports.

\noindent\textbf{Kinetics-600.}
Extending the Kinetics-400 dataset~\cite{Kay2017-gg}, Kinetics-600~\cite{Carreira2018-ly} features videos representing 600 human action classes. The additional videos, sourced from YouTube, broaden the range of depicted actions to include various interpersonal and person-object interactions and individual actions.

\subsection{Implementation details}

We extract RGB frames and resize them to a resolution of $224\times224$. We employ CLIP (with ViT-B/16 visual encoder) as the VLM and GPT-3.5 as the LLM. We do not provide details on training implementation, as our proposed \OurMethod~is inference-only. The sole hyperparameter, the number of frames sampled from the video ($N=16$), is set to align with state-of-the-art methods.
The number of textual descriptors per-class $M$ varies among different classes, as we do not set it a priori and depends on the output of the LLM.

\subsection{Comparative results}
In Tab.~\ref{tab:zs_action_recog_sota}, it can be seen that vanilla CLIP already has good zero-shot performance across the three datasets. It outperforms training-based methods like ER-ZSAR~\cite{chen2021elaborative} and JigsawNet~\cite{qian2022rethinking} without fine-tuning on video data. The remaining training-based methods adapt CLIP by fine-tuning on Kinetics-400. Most of these approaches are supervised, while MAXI~\cite{lin2023match} and LSS~\cite{ranasinghe2024language} perform fine-tuning on an unlabeled video data collection.
With \OurMethod, we eliminate the need for training, enabling direct inference. On UCF101 and HMDB51, our results significantly surpass the CLIP baseline, achieving +6.3\% and +12.8\% Top 1 accuracy, respectively. Additionally, on Kinetics-600, \OurMethod~improves upon the baseline (+6.8/5.3\% Top1/Top5).
 
\begin{table}[!ht]
\centering
\begin{tabular}{p{2.2cm}M{1.3cm}M{2.1cm}M{1.2cm}M{1.4cm}M{1.4cm}M{1.4cm}M{0.7cm}M{0.7cm}} 
\toprule
\textbf{Method}   & \multirow{2}{*}{\textbf{Training}}  & \textbf{Backbone} & \textbf{Frames} &

\textbf{UCF101} & \textbf{HMDB51} & \multicolumn{2}{c}{\textbf{K600}} \\
                                    &              &             &         &    \textbf{Top1}    &   \textbf{Top1}    &   \textbf{Top1} & \textbf{Top5} \\
\midrule

ER-ZSAR~\cite{chen2021elaborative}  & \checkmark & TSM         & 16    &  51.8 & 35.3  & 42.1  & 73.1 \\

JigsawNet~\cite{qian2022rethinking} & \checkmark & R(2+1)D     & 16    &  56.0 & 38.7  & -     & -  \\

ActionCLIP~\cite{wang2021actionclip}       & \checkmark & ViT-B/16    & 32    &  58.3 & 40.8  & 66.7  & 91.6  \\

XCLIP~\cite{ni2022expanding}              & \checkmark & ViT-B/16    & 32    &  72.0 & 44.6  & 65.2  & 86.1  \\

A5~\cite{ju2022prompting}                 & \checkmark & ViT-B/16    & 32    &  69.3 & 44.3  & 55.8  & 81.4 \\

ViFi-CLIP~\cite{rasheed2022fine}    & \checkmark & ViT-B/16    & 32    &  76.8 & 51.3  & 71.2  & 92.2 \\

Text4Vis~\cite{wu2023revisiting}    & \checkmark & ViT-L/14    & 16    &  -    & -     & 68.9  & -\\

MAXI~\cite{lin2023match}            & \checkmark  & ViT-B/16    & 16/32 & 78.2  & 52.3 &  \textbf{71.5}  &  \textbf{92.5}     \\

LSS~\cite{ranasinghe2024language}  & \checkmark  & ViT-B/16    & 8 & 74.2   & 51.4  & -  &  -     \\

\rebuttal{OTI~\cite{zhu2023orthogonal}}& \rebuttal{\checkmark}  & \rebuttal{ViT-B/16}    & \rebuttal{8} & \rebuttal{\textbf{88.3}}   & \rebuttal{\textbf{54.2}}  & -  &  -     \\

\rebuttal{EPK-CLIP\cite{yang2024epk}}  & \rebuttal{\checkmark}  & \rebuttal{ViT-B/16}    & \rebuttal{8} & \rebuttal{75.3}     & \rebuttal{48.7}    & -  &  -     \\

\rebuttal{EPK-ViFi\cite{yang2024epk}}  & \rebuttal{\checkmark}  & \rebuttal{ViT-B/16}    & \rebuttal{8} & \rebuttal{77.7}     & \rebuttal{51.6}    & -  &  - \\

\midrule

CLIP~\cite{radford2021learning}                    & \ding{55}  & ViT-B/16    & 16    &  69.9  & 38.0  & 63.5 & 86.8   \\

\rowcolor{ModelGreen}
\OurMethod                    & \ding{55}  & ViT-B/16    & 16    & 76.2   & 50.8 & 70.3 & 92.1 \\
\bottomrule \noalign{\vskip3mm} 
\end{tabular}
\caption{\textbf{Comparison with state-of-the-art zero-shot action recognition methods.}
We report zero-shot action recognition results on UCF101, HMDB51, and K600. We report Top1 and Top5 accuracy computed on the three official test splits. We also include the backbone used and the number of frames sampled from videos. The green color is \inlineColorbox{ModelGreen}{our method}.}
\label{tab:zs_action_recog_sota}
\end{table}

\subsection{Ablation}\label{sec:ablation_study}
\begin{figure}[!ht]
    \centering
    \includegraphics[width=\textwidth]{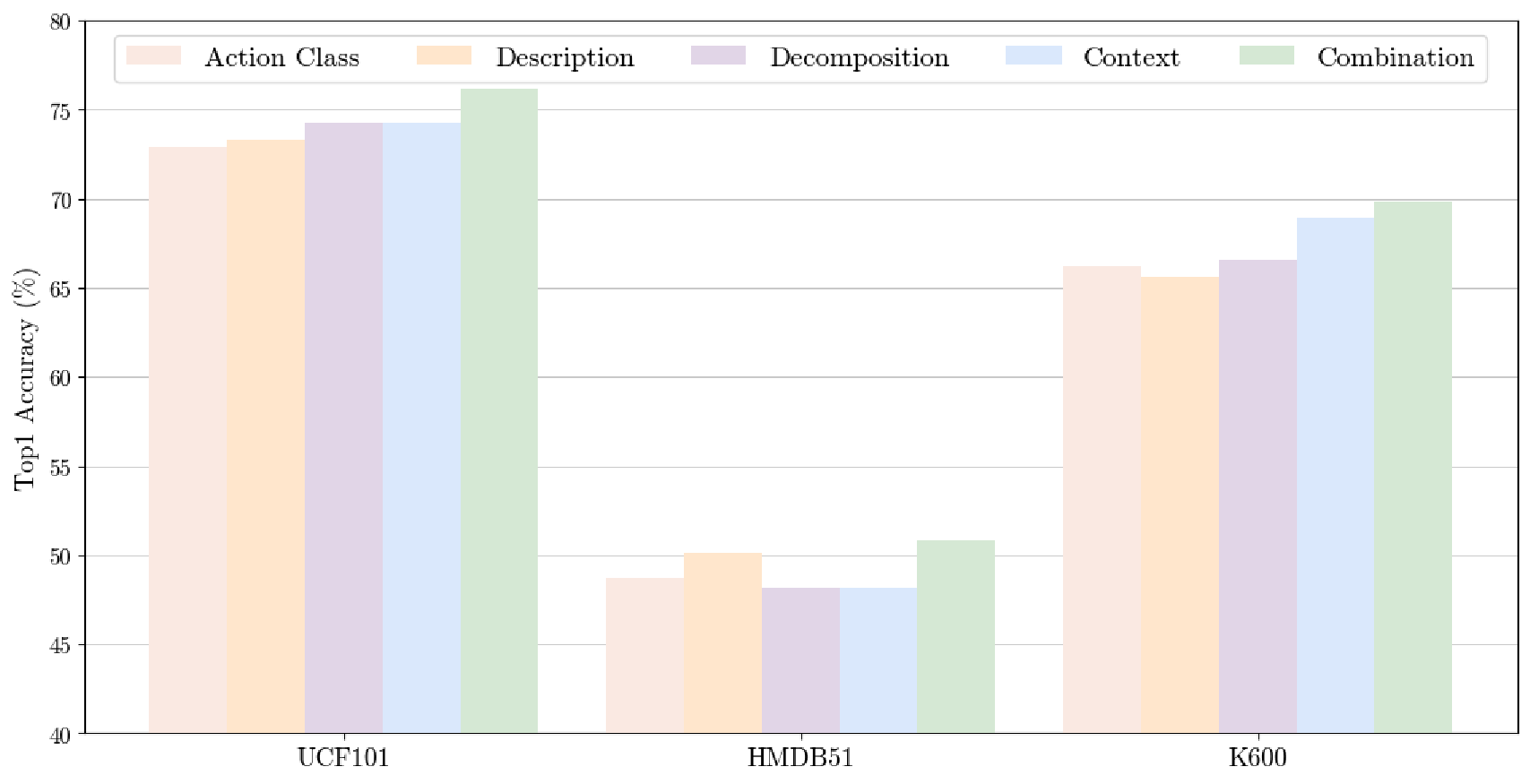}
    \caption{\textbf{Ablation on using the textual descriptors.} We ablate the use of different textual descriptors defined in Sec.~\ref{subsec:descriptors}. We report the Top1 accuracy on the three datasets and use the same color coding as in Sec.~\ref{subsec:descriptors}.}
    \label{fig:ablation_descriptors}
\end{figure}

In this section, we perform ablations of our method to validate our main design choices. We report Top1/Top5 accuracy for all of the datasets considered. 
In the ablation shown in Fig.~\ref{fig:ablation_descriptors}, we evaluate the choice of the textual descriptors used in the proposed methodology, as detailed in Sec.~\ref{subsec:descriptors}.
Our findings indicate that incorporating one of the descriptors usually enhances performance. However, the most substantial improvement is observed when all the descriptors are used in conjunction. Hence, a comprehensive approach furnishes the VLM model with richer linguistic cues, enhancing its zero-shot action recognition accuracy across all benchmarks. In Tab.~\ref{tab:ablation_textprompt}, we depict the results of the ablations of different descriptors, discussed in detail in supplementary material,
the use of additional templates and the choice of prepending the original action class to the obtained textual descriptors. We observe that adding the original class label and using templates enhances the model's accuracy.
\begin{table}[!ht]
\centering
    \begin{tabular}{M{1.8cm}M{1.5cm}M{2cm}M{1cm}M{1cm}M{1cm}M{1cm}M{1cm}M{1cm}}
         \toprule
         \textbf{Template} & \textbf{Class} & \textbf{Backbone} & \multicolumn{2}{c}{\bf UCF101} & \multicolumn{2}{c}{\bf HMDB51} & \multicolumn{2}{c}{\bf K600} \\
                     &    & & \bf Top1 & \bf Top5 & \bf Top1 & \bf Top5 & \bf Top1 & \bf Top5 \\  \noalign{\vskip1mm}

\toprule
\ding{55} & \ding{55} & ViT-B/32 & 68.9 & 93.1 & 43.6 & 73.9 & 61.1 & 86.9 \\
\checkmark & \ding{55} & ViT-B/32 & 68.7 & 92.8 & 45.8 & 74.1 & 61.5 & 87.4 \\
\ding{55} & \checkmark & ViT-B/32 & 72.2 & 94.0 & 47.5 & 76.9 & 66.9 & 89.5 \\
\checkmark & \checkmark & ViT-B/32 & 72.6 & 94.6 & 48.0 & 78.5 & 67.0 & 90.0 \\
\midrule
\ding{55} & \ding{55} & ViT-B/16 & 70.1 & 94.0 & 44.1 & 75.6 & 66.0 & 90.5 \\
\checkmark & \ding{55} & ViT-B/16 & 72.6 & 94.6 & 48.4 & 77.0 & 66.2 & 90.1 \\
\ding{55} & \checkmark & ViT-B/16 & 75.8 & 95.9 & 49.0 & 81.2 & 70.2 & 92.2 \\
\rowcolor{ModelGreen}
\checkmark & \checkmark & ViT-B/16 & \textbf{76.2} & \textbf{96.3} & \textbf{50.8} & \textbf{82.0} & \textbf{70.3} & \textbf{92.3} \\
        \bottomrule \noalign{\vskip2mm}
    \end{tabular}
    \caption{\textbf{Ablation on constructing the textual prompts.} We ablate using templates and prepending the action class after descriptor generation. Results are reported for both ViT-B/32 and ViT-B/16 visual backbones. Green is \inlineColorbox{ModelGreen}{our configuration}.}\label{tab:ablation_textprompt}
\end{table}

In addition, in Tab.~\ref{tab:ablation_backbone}, we assess the choice of the visual backbone $\mathcal{E}_I$ and the number of sampled frames $N$.  
We observe a significant gain with ViT-B/16 compared to ViT-B/32, and ViT-B/16 is also the backbone commonly employed by other competitors. Additionally, our method exhibits low sensitivity to the number of sampled frames for both backbones, whether 16 or 32. As a result, we adopt the configuration with 16 sampled frames as our final choice to have a fair comparison with most competitors who also utilize this setting.

\begin{table}[ht]
\centering
    \begin{tabular}{p{2cm}M{0.5cm}M{1.25cm}M{1.25cm}M{1.25cm}M{1.25cm}M{1.25cm}M{1.25cm}}
        \toprule
        \bf Backbone & \bf N  & \multicolumn{2}{c}{\textbf{UCF101}} & \multicolumn{2}{c}{\textbf{HMDB51}} & \multicolumn{2}{c}{\textbf{K600}} \\
                        &  & \textbf{Top1} & \textbf{Top5} & \textbf{Top1} & \textbf{Top5} & \textbf{Top1} & \textbf{Top5} \\ 
        \midrule
    
ViT-B/32 & 32 & 72.5 & 94.7 & 48.5 & 78.2 & 67.0 &  90.0 \\
ViT-B/32 & 16 & 72.2 & 94.0 & 47.5 & 76.9 & 66.8  & 89.9 \\
ViT-B/16 & 32 & 76.4 & 96.5 & 50.8 & 82.2 & 70.5 &  92.2 \\
\rowcolor{ModelGreen}
ViT-B/16 & 16 & 76.2 & 96.3 & 50.8 & 82.0 & 70.3 & 92.3 \\
        
         \noalign{\vskip1mm}
        \bottomrule \noalign{\vskip2mm}
    \end{tabular}
    \caption{\textbf{Ablation on the backbone used and the frame sampling N.} Green is \inlineColorbox{ModelGreen}{our configuration}.}\label{tab:ablation_backbone}
\end{table}

\rebuttal{Lastly, we ablate the different LLMs to determine the robustness of the method related to the generation of action descriptors. Thus, we re-evaluate our method using different LLMs on the HMBD dataset.}
\begin{table}[!ht]
\centering
    \begin{tabular}{p{1.8cm}M{1.2cm}M{1.2cm}M{1.2cm}M{1.2cm}M{1.2cm}M{1.2cm}M{1.2cm}M{1.2cm}}
        \toprule
        
        \rebuttal{\bf LLM} & \multicolumn{2}{c}{\rebuttal{\textbf{Description}}} & \multicolumn{2}{c}{\rebuttal{\textbf{Decomposition}}} & \multicolumn{2}{c}{\rebuttal{\textbf{Context}}} & \multicolumn{2}{c}{\rebuttal{\textbf{Combination}}} \\

       &  \rebuttal{\textbf{Top1}}   & \rebuttal{\textbf{Top5}} & \rebuttal{\textbf{Top1}} & \rebuttal{\textbf{Top5}} & \rebuttal{\textbf{Top1}} & \rebuttal{\textbf{Top5}} & \rebuttal{\textbf{Top1}} & \rebuttal{\textbf{Top5}} \\ 
        \midrule
        \rowcolor{ModelGreen}
        \rebuttal{GPT-3.5}  & \rebuttal{50.1} & \rebuttal{79.2} &\rebuttal{48.2} & \rebuttal{81.1} & \rebuttal{48.2} & \rebuttal{81.1}  & \rebuttal{50.8} & \rebuttal{82.0} \\
        \rebuttal{GPT-4o~\cite{openai2024chatgpt}} & \rebuttal{51.4} & \rebuttal{81.3} &\rebuttal{49.7} & \rebuttal{79.8} & \rebuttal{49.9} & \rebuttal{80.1}  & \rebuttal{50.8} & \rebuttal{82.3} \\

        \rebuttal{Llama3~\cite{meta2024introducing}} & \rebuttal{49.5} & \rebuttal{80.2} &\rebuttal{49.1} & \rebuttal{77.9} & \rebuttal{49.5} & \rebuttal{79.4}  & \rebuttal{49.5} & \rebuttal{80.1} \\

        \rebuttal{Mistral~\cite{jiang2023mistral}} & \rebuttal{47.8} & \rebuttal{78.3} &\rebuttal{47.4} & \rebuttal{78.9} & \rebuttal{47.8} & \rebuttal{79.8}  & \rebuttal{45.2} & \rebuttal{74.8} \\
        
         \noalign{\vskip1mm}
        \bottomrule \noalign{\vskip2mm}
    \end{tabular}
    \caption{\rebuttal{\textbf{Ablation on the LLM used to generate prompts.} Green is \inlineColorbox{ModelGreen}{our configuration}.}
    }\label{tab:ablation_llm}
\end{table}
\rebuttal{
Tab.~\ref{tab:ablation_llm} revealed that our method is robust, with only minor performance differences across different LLMs. Although advanced models like GPT-4o offer slight improvements, our method remains effective regardless of the model used.
This showcases the method's reliability and adaptability to various LLMs with varying capacities.
We maintain the use of GPT-3.5 as it offers a cost-effective alternative to GPT-4o, ensuring the method remains accessible without sacrificing significant performance.}

In conclusion, our experiments, which combine all forms of text augmentation —including label, description, decomposition, and context— significantly when templates and class label conditioning are applied, demonstrate a cumulative improvement in performance.

\section{Limitations}
While our approach for generating textual action descriptors provides an automated pipeline for capturing various aspects of action classes, it may be suboptimal for more temporally fine-grained or very atomic actions that cannot be decomposed into distinct steps.
Additionally, our method may encounter limitations when dealing with actions that exhibit less association with objects or are highly variable in context. 
Actions with weaker object associations may benefit less from the generated textual descriptors.
Similarly, actions that vary widely in context may result in descriptors that fail to capture the diverse contexts in which they occur. 
Addressing these limitations can lead to more advanced models for language-driven action recognition.

\section{Conclusion}
This work tackles the challenging problem of zero-shot video action recognition. We propose \OurMethod, a training-free approach that generates rich textual descriptors for the action class labels and then performs zero-shot prediction using the obtained descriptors. 
Despite its simplicity, \OurMethod~outperforms baseline models and rivals training-based methods in the task of zero-shot action recognition, all without the need for in-domain training.

While our method was primarily evaluated on action recognition, its applicability can extend to more challenging tasks on untrimmed videos, such as Temporal Action Localization or Action Segmentation. By leveraging textual descriptors to bridge the semantic gap between action labels and visual content, our approach promises to tackle broader video understanding tasks beyond mere classification. Future research efforts should explore the adaptation and extension of \OurMethod~to address these more complex video analysis tasks.

\subsection*{Acknowledgements}
We acknowledge the PRECRISIS project, funded by the EU  Internal Security Fund (ISFP-2022-TFI-AG-PROTECT-02-101100539), MUR PNRR project FAIR - Future AI Research  (PE00000013), funded by NextGeneration EU and the CINECA award under the ISCRA initiative for the availability of high-performance computing resources and support. We also thank the Deep Learning Lab of the ProM Facility for the GPU time.


%
%
%

{\small
\bibliographystyle{splncs04}
\bibliography{mybibliography}
}


\end{document}